# A Method for Restoring the Training Set Distribution in an Image Classifier

## An Extension of the Classifier-To-Generator Method

**Alexey Chaplygin & Joshua Chacksfield**
Data Science Research Group, Department of Analytics and Business Intelligence, PVH Corp.
{alexey.chapligin, chacksfieldj}@gmail.com

**ABSTRACT**

*Convolutional Neural Networks are a well-known staple of modern image classification. However, it can be difficult to assess the quality and robustness of such models. Deep models are known to perform well on a given training and estimation set, but can easily be fooled by data that is specifically generated for the purpose. It has been shown that one can produce an artificial example that does not represent the desired class, but activates the network in the desired way. This paper describes a new way of reconstructing a sample from the training set distribution of an image classifier without deep knowledge about the underlying distribution. This enables access to the elements of images that most influence the decision of a convolutional network and to extract meaningful information about the training distribution.*

## 1. Introduction

While convolutional neural networks have been shown to have great performance on image classification tasks [1], these deep Networks are generally seen as "black boxes" and are believed to be hard to analyse after training. In recent years a number of techniques derived from adversarial examples were invented which are capable of fooling such image classifiers. Creating an adversarial example is an easy task, and can produce comically wrong results (See Figure 1) for text captioning. In the realm of image classification it has been shown that images can be generated that are classified with a high confidence but are unrecognisable to humans [2].

The miss categorisation of images can cause serious issues in commercially available classifiers, leaving them open to attacks and potential vulnerabilities. The question as to which features influence the output of a convolutional network is often posed and is difficult to answer due to the lack of knowledge and understanding of the internals of convolutional networks.

Several techniques have been developed that provide insights into the internals of trained models. Such approaches range from simply slicing the convolutional layers of a network and visualising the neurons to constructions called grad cams or guided grad cams [3]. However all of these techniques require samples from the distribution upon which the model was trained and do not show the limits of each trained classifier. Acquiring a sample of training data can be difficult in cases where the nature of a classifier is barely known and is not trained on a standard open dataset such as ImageNet.

There are multiple cases where the nature of the classifier might be unknown, for example digital locks that use iris, retina or facial recognition. Another potential use for the described method is to determine the distribution of data upon which an image classifier was trained and access whether the classifier could be applied to a particular problem.

## 2. Related Work

There has been some other research conducted, independent of this article, which aimed to provide a view into the training distribution. Anonymous authors detailed an investigation titled "Classifier-to-Generator Attack: Estimation of Training Data Distribution from Classifier", which generally coincides with the findings in this paper, that the training distribution can be reconstructed from a classifier.

However, the main difference between classifier-to-generator attack approach and the approach described below is that this approach does not require structural similarity between the substrate and distribution to restore an auxiliary dataset. The aforementioned paper used very similar datasets for both, which implies a good understanding of the distribution of the training set of the classifier. Another difference is the use of Conditional GAN (Conditional Generative Adversarial Nets) in this paper as opposed to classic GAN. This was in order to have a substrate image as a starting point of the reconstruction. When using a classic GAN [4] the model was unable to restore a distribution of unknown nature and wasn't applicable in our case (Figure 2).

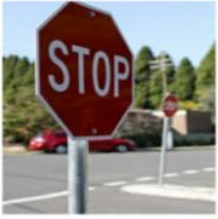
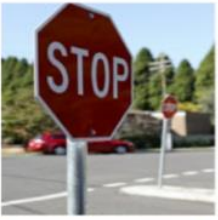
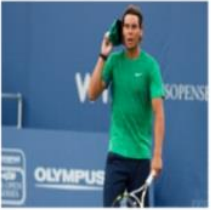
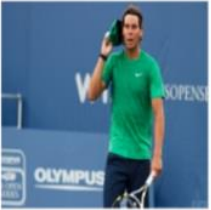

**Figure 1.** Adversarial examples crafted by Show-and-Fool using the targeted caption method **[12]**. Examples showing how alterations to images can cause gross 'misclassification' of images. Top of each pair shows the original image with the correctly generated text. Bottom of each pair contains the crafted example that results in highly different caption to be generated.

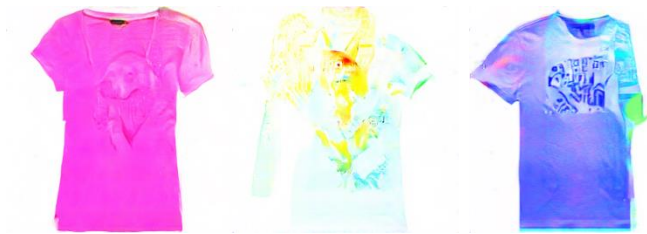

**Figure 2.** These are some of results which best showed the required categories, when using only classic GAN as opposed to while using Conditional GAN.

**Note:**
This paper is the result of a project executed for PVH Europe where the business need was to try to generate body-wear with prints/patterns that have a definite association with some defined category, but does not repeat in the exact full details of the object. In such a way interesting prints that remind us of objects from nature can be created, if to use a blank piece of clothes as a substrate.

For a better understanding of the results and the way of repeating the experiment a practitioners approach was used rather than academic.

## 3. Methodology

In this example a VGG16 network was used, pre-trained on 1000 classes, but this method could be applied to any image classifier. The 1000 classes were filtered down to 5 classes that were to be reconstructed and are represented from here on as the target vector

$$\vec{t} \in \{0,1\}^n = \begin{pmatrix} t_1 \\ t_2 \\ \vdots \\ t_n \end{pmatrix}$$

where $n = 5$. During training only one class was ever positive per batch of $m$ images with the active class selected randomly with equal probability for each batch of images.

To generate the images a Conditional GAN was used. The generator, $G$ (Figure 4) was constructed such that, for each convolution layer, the input was padded and filters of size 4 were applied with a stride of size 2. Lastly a PReLU (Parameterised Leaky ReLU function initialised to 0.2) was applied to improve model fitting [5]. The deconvolution layers were constructed in a similar manner, using filters of size 4, applied with a stride of size 2, also applying PReLU for all but the last layer for which tanh function was used.

A random substrate image of size of 512x512 was convolved into feature maps of sizes between 16x16 and 128x128. The input to the deconvolutional layer was a "Tiled Class Input" created by a function,

$$f(t_i), f: \{0,1\} \to \mathbb{R}^{8,8}$$

where $t_i$ represents the elements of $\vec{t}$ and

$$f(x_i) = \begin{cases} \{0\}^{8,8} & (x_i = 0) \\ A \sim \big(\mathcal{N}(1,2) + \mathcal{N}(-1,2)\big)^{8,8} & (x_i = 1) \end{cases}$$

such that $f$ is a function applied to each element of a vector $\vec{x}$ that returns an $8 \times 8$, **0** matrix if the input element $x_i = 0$. Otherwise if $x_i = 1$ the result was a random matrix where each element was selected from the union of two distributions, $\mathcal{N}(1,2)$ and $\mathcal{N}(-1,2)$ also of size $8 \times 8$. The result is a 3 dimensional matrix $(n, 8, 8)$ where $n$ is the size of vector $\vec{x}$.

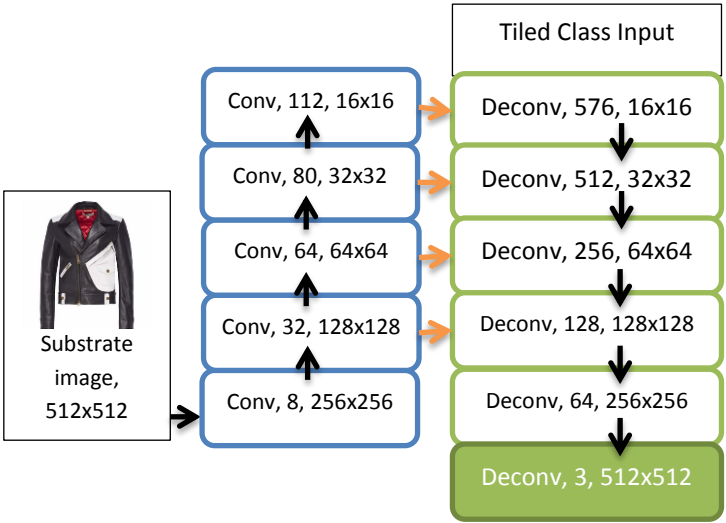

**Figure 4.** Architecture of the generator $G$ used in the conditional GAN.

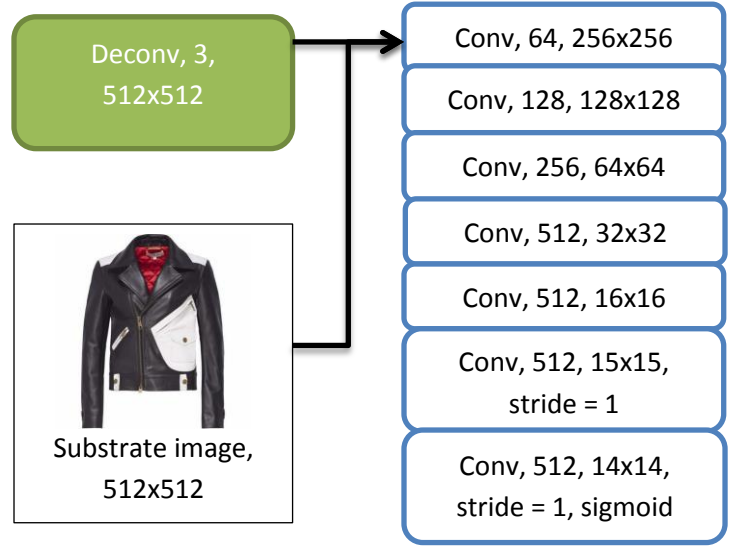

**Figure 3.** Architecture of the discriminative part $D$ of the conditional GAN, used in conjunction with the generator $G$.

This construct was put to the deconvolutional layers along with a U-Net structure where the feature maps of the convolutional part of $G$ are also concatenated to the deconvolutional layers outputs and fed to the next layer [6] to reconstruct an image of size 512x512 (Figure 4). During training Weight Normalization was applied together with Parametric Leaky ReLU for faster convergence and improved stability [7].

The discriminative part $D$ of our Condition GAN was a basic CNN based on PatchGAN architecture [8]. The substrate image and output of $G$ are supplied to the network as inputs (Figure 3).

Multiple loss functions were used to construct the objective function for this project. There are four parts to the loss function used, firstly a canonical GAN loss, for which $x$ is defined as the observed image, $z$ the noise component and $y$ the output image, are all passed to the network.

$$L_{cGAN} = \mathbb{E}_{x,y}[\log D(x,y)] + \mathbb{E}_{x,z}\left[\log\left(1 - D(x, G(x,z))\right)\right]$$

where $G$ tries to minimise this against adversarial $D$ that tries to maximise, i.e. $G^* = \arg\min_G \max_D L_{cGAN}(G, D)$ [9]. Secondly, a masking loss,

$$L_m = \mathbf{1} - M(G)$$

where $M(G)$ is the masked output of the model and $\mathbf{1}$ is the unit matrix. For the situation in which the model was to be applied it was desired to keep the background of the image as white as possible, which was accomplished by penalizing any output other than white.

A pre-trained VGG16 classifier was used to penalize output (as seen in Figure 5) in the following way. Let $C$ be the set of all square crops of size 224 pixels in the image generated by the GAN. Let $C_1, C_2, C_3 \in C$ s.t. $C_1 \neq C_2 \neq C_3$ be three distinct crops selected randomly with equal probability. Let

$$\vec{v} \in [0,1]^n = \begin{pmatrix} v_1 \\ v_2 \\ \vdots \\ v_n \end{pmatrix}$$

be the vector of outputs from the VGG classifier for n classes. After processing the three random crops and resized image a vector with the probability of each class was resolved, $\vec{c}_1$, $\vec{c}_2$, $\vec{c}_3$ and $\vec{c}_r$ respectively.

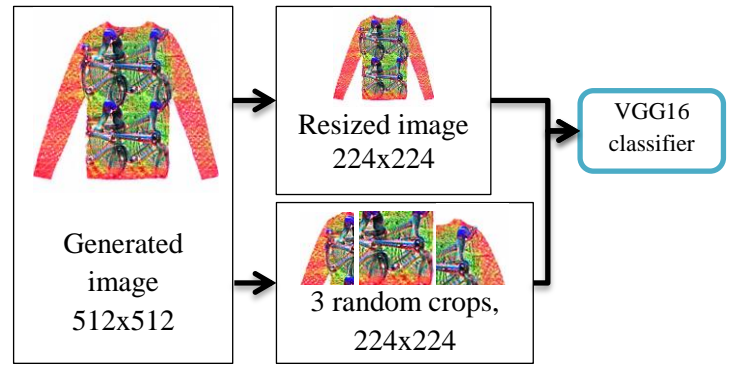

**Figure 5.** For each generated image a loss was applied to determine whether the required categories were present on the image. Classification was performed in two different ways, firstly on the whole image resized and secondly to three random crops. The classifications were then used in the loss $L_{VGG}$.

The difference from the target, $\vec{t}$ for positive classes was calculated upon the whole resized image only as

$$P(\vec{v}) = |\vec{t} - \vec{v}| \odot \vec{t}$$

and the difference for negative classes was calculated upon the three random crops as

$$N(\vec{v}) = |\vec{t} - \vec{v}| \odot (\mathbf{1} - \vec{t})$$

where, $c = a \odot b$ is defined as the element wise multiplication of two vectors or $c_i = a_i b_i$ for 'element' $i \in \mathbb{Z}, n \geq i > 0$ in each vector.

The final loss constraint for the classifiers category being present on the generated image was constructed from two parts, the positive classes appearing in the entire picture and the negative classes in the random crops. For each a simple summation of a linear and logarithmic term is applied to a vector $\vec{x} \in [0,1]^n$, calculated as

$$L(\vec{x}) = \frac{1}{n} \sum_i^n (x_i - \log(1 - x_i))$$

So for each of the crops the loss term for appearance of negative categories is

$$L_n = \max(\{L \circ N(\vec{c}) : \vec{c} = \vec{c}_1, \vec{c}_2, \vec{c}_2 \})$$

and for the require class in the resized image

$$L_p = L \circ P(\vec{c}_r)$$

which results in the total loss for the categories being the combination of the positive and negative classes.

$$L_{VGG} = L_p + L_n$$

In that way we prevent the situation when all categories are being merged together and shown simultaneously. With a certain frequency we require a model to generate image free of any ImageNet category.

An extra addition to the loss was a term named the substrate loss. This was a term that encouraged the model to reconstruct an output similar to the substrate image. It is calculated as,

$$L_s = |T - G| - \log\left(1 - \left|\frac{T - G}{2}\right|^2\right)$$

where $T$ is the target image and $G$ is the generated image. Practically good results were achieved with the following ratio of loss terms:

$$L = 3L_{cGAN} + 10L_m + 50L_{VGG} + 150L_s$$

An Adam optimizer [10] was used to minimize the target function with a *learning rate* of .0002 and *beta1* of .5. With a batch size of between 1-3 images and number of steps of 200000 training was accomplished within ~5-15 hours on one NVidia Titan Xp GPU.

## Dataset

The auxiliary dataset consists of ~2800 images of body-wear from Tommy Hilfiger. The data was augmented by combining colour swaps (swapping colour axes), colour shifts (replacing all values of a random colour axes to 255 if information loss does not occur and number of unique colours remains the same) and horizontal flips.

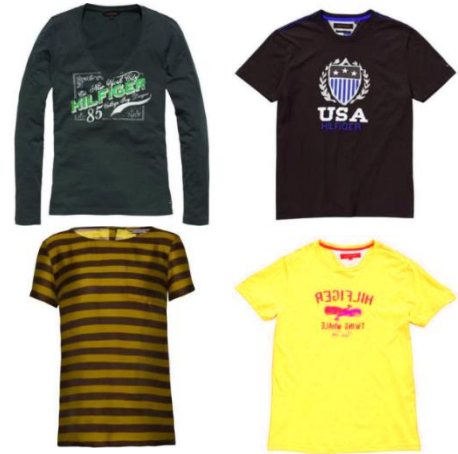

**Figure 6.** Example substrate images

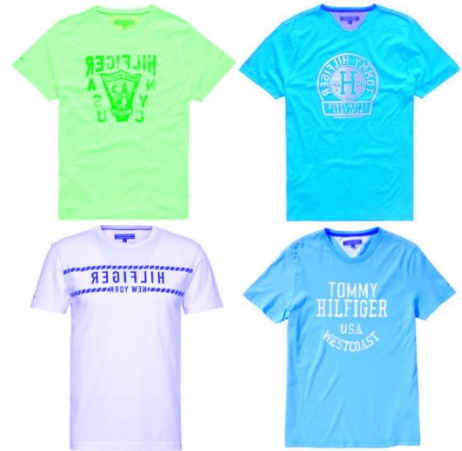

**Figure 7.** Example colour shifted substrates

# 4. Results

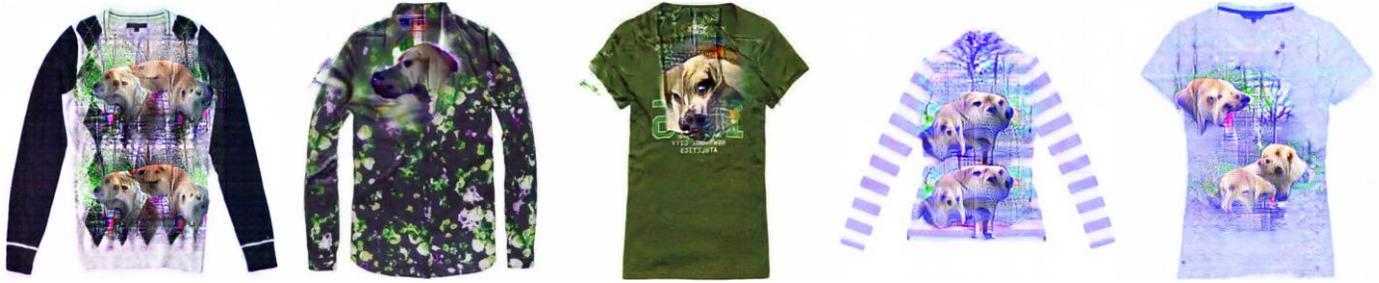

Labrador (green grass/leaves and tree-like structures behind are also considered as a "Labrador")

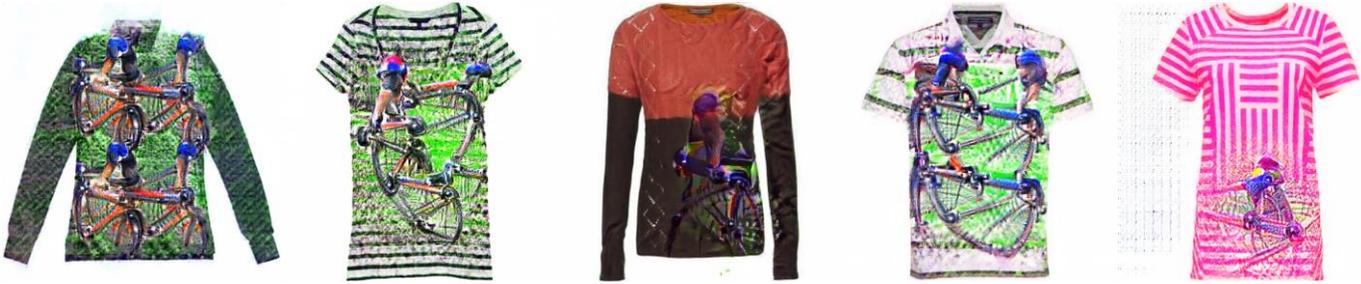

Tandem-bike (silhouette of a person/body parts)

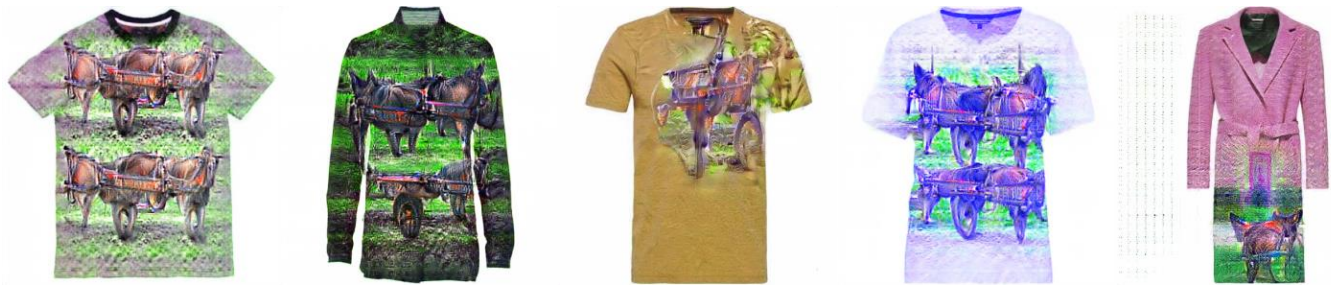

Horse-cart (with grass and ground)

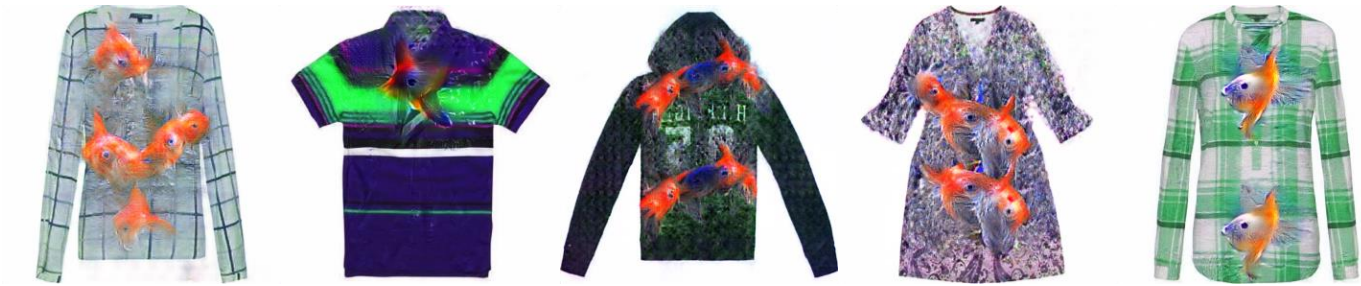

Gold-fish (pretty clearly identified)

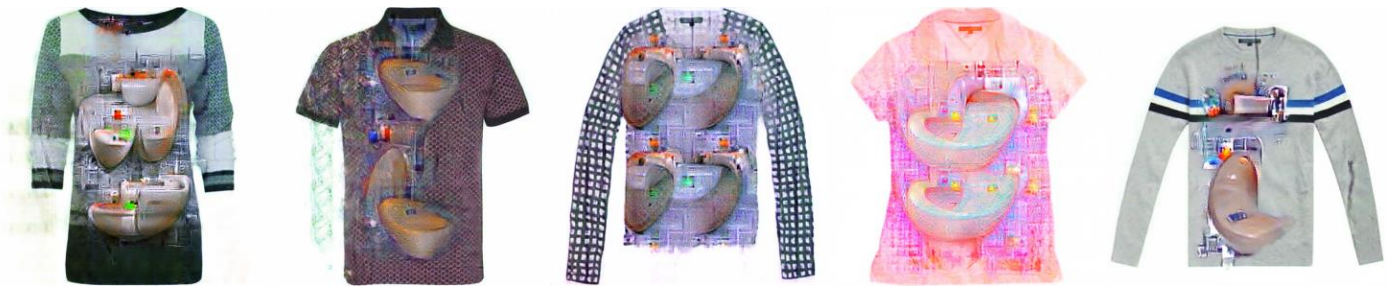

Toilet seat (with pipes)

Providing gray-scale image to thee VGG part of the model gave even better results due to the decreased adversarial effect of trying to generate an object that looks like a body wear:

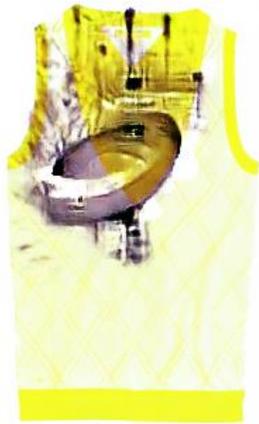
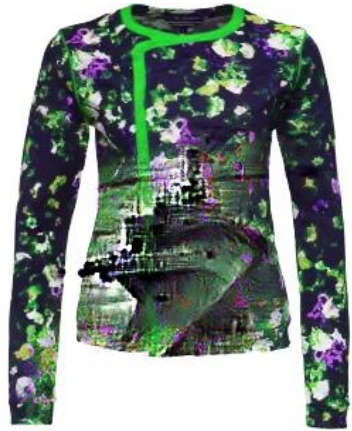
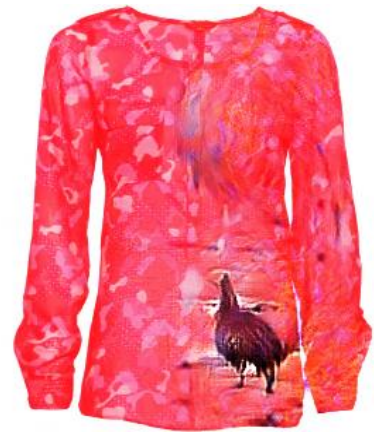

**Figure 8.** Toilet seat (with pipes)  **Figure 9.** Aircraft carrier (water/sky also considered as an aircraft carrier)  **Figure 10.** Cock

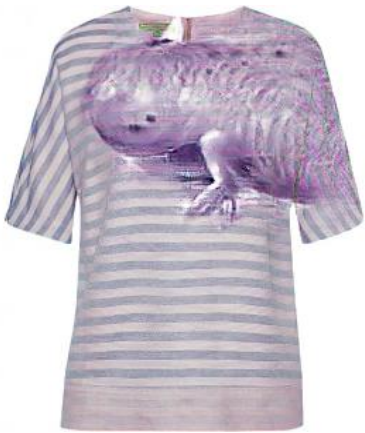
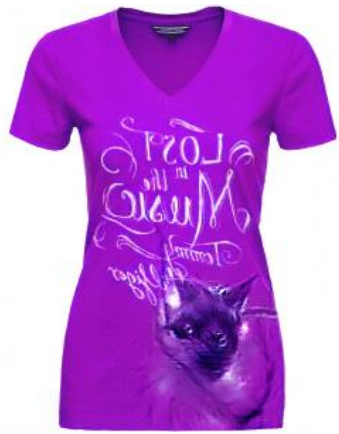
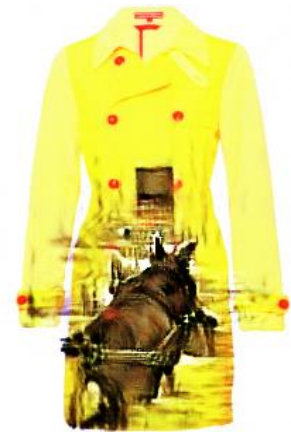

**Figure 11.** Salamander  **Figure 12.** Siamese cat  **Figure 13.** Horse-cart

Different ratios of the optimiser's targets combined with mixing multiple categories can give interesting results:

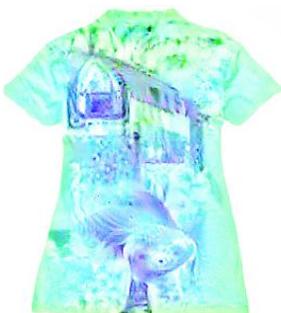
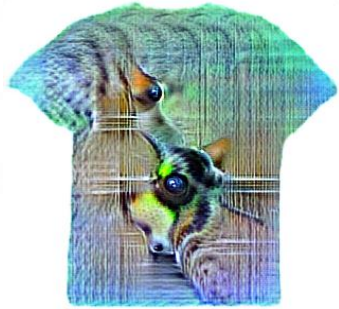
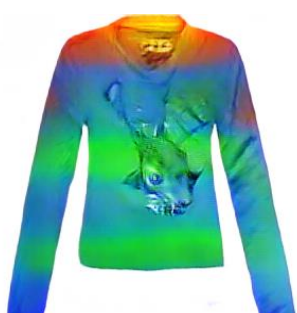
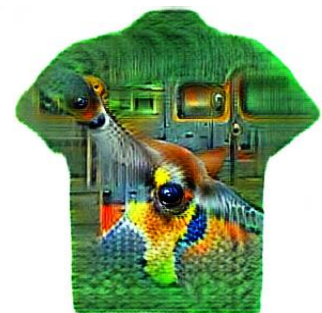

**Figure 14.** Train + salamander  **Figure 15.** Dog + cat  **Figure 16.** Fish + rabbit  **Figure 17.** Dog + train + goldfish

Such generators using the architecture described in this paper were able to restore up to 24 categories, with up to 3 categories mixed together at once. When using a higher number of categories the generator failed to effectively reconstruct all of the categories and objects became unrecognizable mixing details of one category with those from another.

In an extension to the method described above, an approach similar to GradCam [11] was used to mask the regions which activated the VGG classifier the most. This alteration provided visually better separation between substrate image and the generated picture of desired category. However, it was not applicable in the scope of the original project for which a smooth merge between the substrate and generated category was required. Such a masking is definitely advised for achieving better results when reconstructing the training distribution.

## 5. Conclusion

When applying such architectures over a classifier, a sample of the training distribution can be reconstructed. This approach opens up a host of possible vulnerabilities for image recognition models, some with serious consequences. Examples include any system that is used for authorisation purposes could be exploited or a model trained on sensitive or private data could be opened up for public access.

At the same time the described approach can be used to visualize the capabilities of image classifiers and analyse redundancy within the models. On the samples above it can be clearly seen, that VGG16 is being activated on artefacts that do not have a direct relation to the predicted category. The *"Labrador"* category also reproduced trees and in the *"Horse-Cart"* category, grass and other details around the horse can be observed. This shows the redundancy of the classifier itself and/or not optimal training set.

We believe that the model is able to reconstruct the training distribution when using an auxiliary dataset that is far from the training distribution, due to the fact, that Conditional GAN is used. The conditional GAN only determines whether or not the generated image *looks* like a photo of an actual object from a "real" world. However, this is also a restriction of the approach, the auxiliary dataset should consist of images containing the style we would like to inherit in the generated object. It was known that VGG was trained on photos and so it was possible to reconstruct samples from the training set of VGG, using an auxiliary dataset also containing photos.

Investigation into the use of auxiliary datasets with different distributions and using different classifiers is an avenue of further research, which could bring greater understanding to this problem.

## 6. Supplementary Material

The architecture was extended to be able to output images of size of 1024x1024 and optimised the objective function in such a way, that ImageNet category can be clearly seen on each generated picture. However, 100% confidence of the image classifier was not required so as to balance the $L_{VGG}$ and $L_{GAN}$ loss which resulted in more visually pleasing patterns. It was also required that any random crop should contain a wanted category, forcing a model to generate more pattern-like outputs rather than a picture of some certain object.

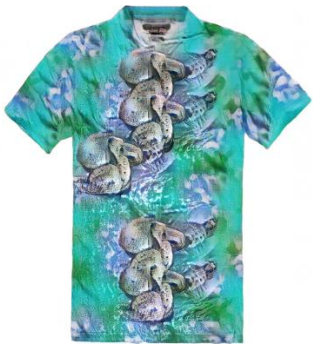
**Figure 18.** Snake

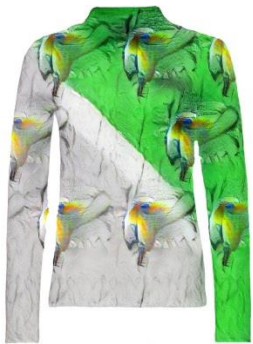
**Figure 19.** Parrot

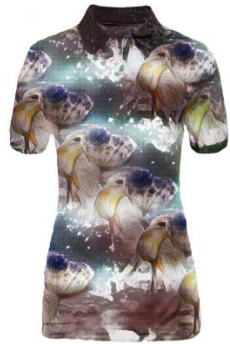
**Figure 20.** Turtle

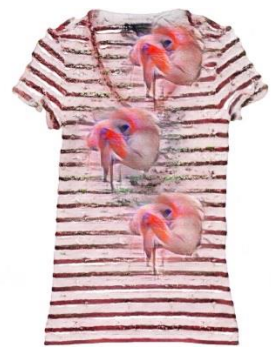
**Figure 21.** Flamingo

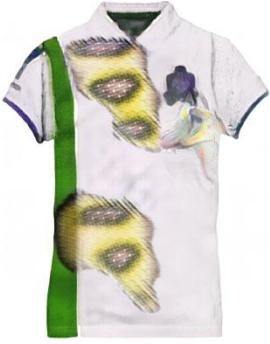
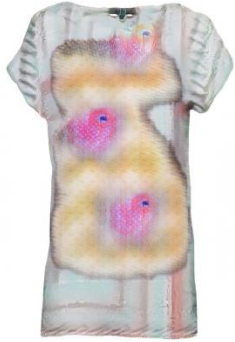
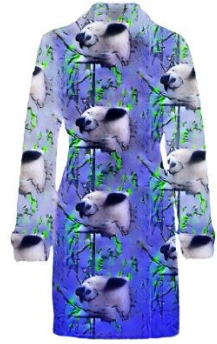
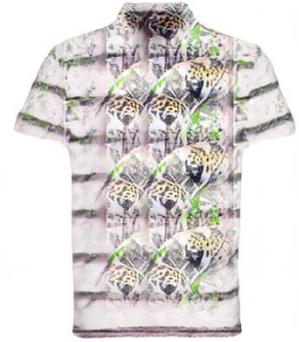

**Figure 22.** Bee

**Figure 23.** Hamster

**Figure 24.** Panda (even without requirement of 100% confidence tree is clearly seen by VGG16 as part of "panda" category)

**Figure 25.** Jaguar

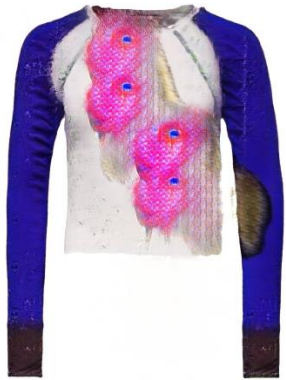
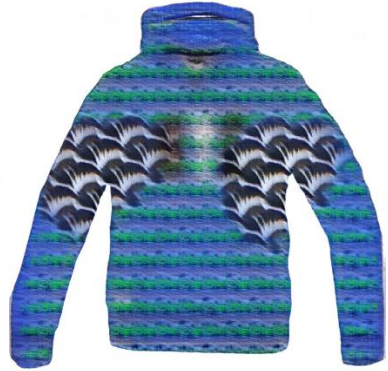

**Figure 26.** Cock

**Figure 27.** Zebra